\documentclass[letterpaper]{article} 
\usepackage{aaai2026}
\usepackage{times}  
\usepackage{helvet}  
\usepackage{courier}  
\usepackage[hyphens]{url}  
\usepackage{graphicx} 
\usepackage{natbib}  
\usepackage{caption} 
\usepackage{algorithm}
\usepackage{algorithmic}
\usepackage{amsmath}
\usepackage{amssymb}
\usepackage{booktabs}
\usepackage{multirow}
\usepackage{amsmath} 
\usepackage{amssymb}
\usepackage{xcolor}
\usepackage{newfloat}
\usepackage{listings}
\DeclareCaptionStyle{ruled}{labelfont=normalfont,labelsep=colon,strut=off} 
\floatstyle{ruled}
\newfloat{listing}{tb}{lst}{}
\floatname{listing}{Listing}
\nocopyright 

\title{SPJFNet: Self-Mining Prior-Guided Joint Frequency Enhancement for Ultra-Efficient Dark Image Restoration}
\author{
    Tongshun Zhang\textsuperscript{\rm 1,\rm 2},
    Pingping Liu\textsuperscript{\rm 1,\rm 2}\thanks{Corresponding author},
    Zijian Zhang\textsuperscript{\rm 1,\rm 2},
    Qiuzhan Zhou\textsuperscript{\rm 3}
}
\affiliations{
    \textsuperscript{\rm 1} College of Computer Science and Technology, Jilin University\\
    \textsuperscript{\rm 2} Key Laboratory of Symbolic Computation and Knowledge Engineering of Ministry of Education, Jilin University\\
    \textsuperscript{\rm 3} College of Communication Engineering, Jilin University\\

    tszhang23@mails.jlu.edu.cn,\{liupp, zhangzijian, zhouqz\}@jlu.edu.cn
}

\usepackage{bibentry}

\begin{document}

\maketitle

\begin{abstract}
Current dark image restoration methods suffer from severe efficiency bottlenecks, primarily stemming from: (1) computational burden and error correction costs associated with reliance on external priors (manual or cross-modal); (2) redundant operations in complex multi-stage enhancement pipelines; and (3) indiscriminate processing across frequency components in frequency-domain methods, leading to excessive global computational demands. To address these challenges, we propose an Efficient \textbf{S}elf-Mining \textbf{P}rior-Guided \textbf{J}oint \textbf{F}requency Enhancement Network (SPJFNet). 
Specifically, we first introduce a Self-Mining Guidance Module (SMGM) that generates lightweight endogenous guidance directly from the network, eliminating dependence on external priors and thereby bypassing error correction overhead while improving inference speed. Second, through meticulous analysis of different frequency domain characteristics, we reconstruct and compress multi-level operation chains into a single efficient operation via lossless wavelet decomposition and joint Fourier-based advantageous frequency enhancement, significantly reducing parameters. Building upon this foundation, we propose a Dual-Frequency Guidance Framework (DFGF) that strategically deploys specialized high/low frequency branches (wavelet-domain high-frequency enhancement and Fourier-domain low-frequency restoration), decoupling frequency processing to substantially reduce computational complexity. 
Rigorous evaluation across multiple benchmarks demonstrates that SPJFNet not only surpasses state-of-the-art performance but also achieves significant efficiency improvements, substantially reducing model complexity and computational overhead. Code is available at https://github.com/bywlzts/SPJFNet.
\end{abstract}


\section{Introduction}
\label{sec:introduction}

\begin{figure}[htbp]
    \centering
    \includegraphics[width=0.45\textwidth]{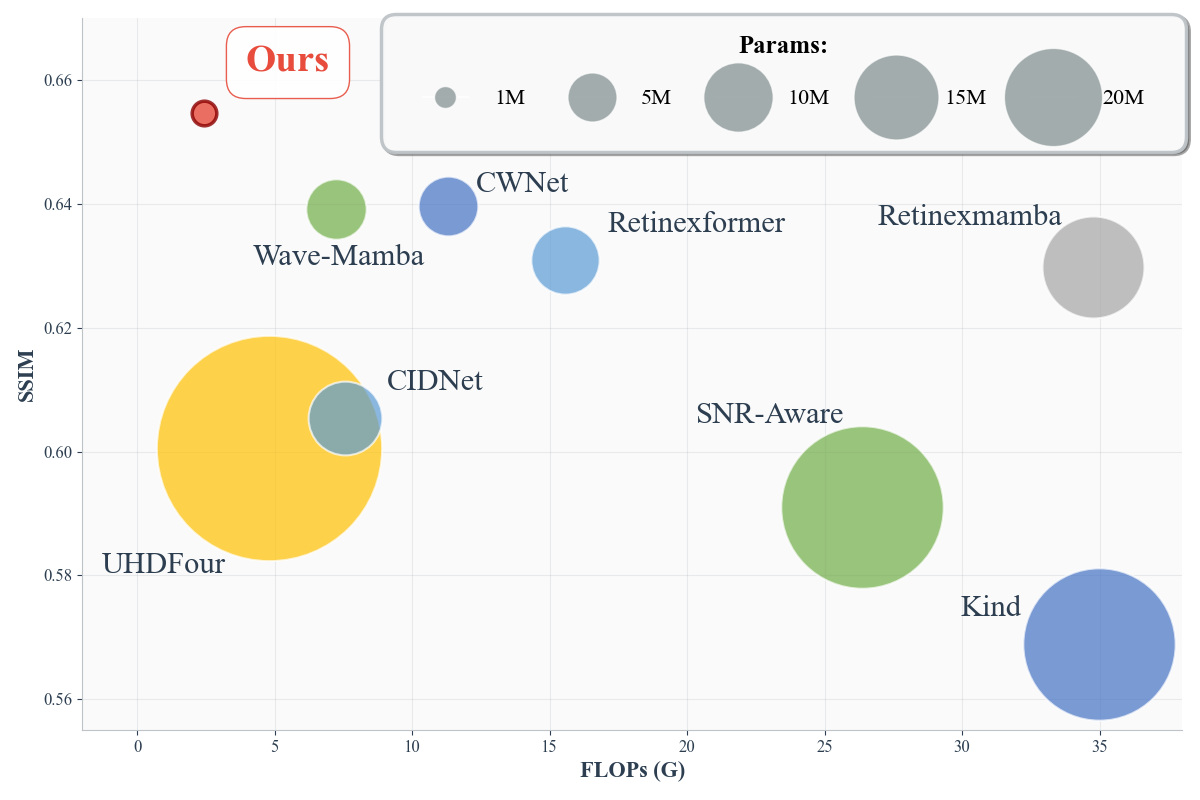}
    \vspace{-0.35cm}
    \caption{SSIM vs. Computation overhead. Our method achieves superior performance with minimal computational cost on the LSRW-Huawei dataset.}
    \vspace{-0.4cm}
    \label{fig:ssim_computation_overhead}
\end{figure}

Images taken in dark conditions often suffer from insufficient illumination, degrading visual quality and hindering downstream vision applications like autonomous driving and object detection~\cite{du2024boosting,li2024light}. With the rise of deep learning, learning-based methods~\cite{retinexformer, zhang2025cwnet, gu2025improving} have surpassed traditional approaches but often come with high computational complexity. Given the limited cues in extremely dark images, many methods~\cite{wu2023learning, zhang2024dmfourllie, zou2024vqcnir} introduce pre-trained models or additional prediction networks to generate modal priors. However, pretrained models and the predictive network are likely to suffer from generalization issues, resulting in accumulated errors and high computational overhead.

\begin{figure*}[t]
    \centering
    \includegraphics[width=1\linewidth]{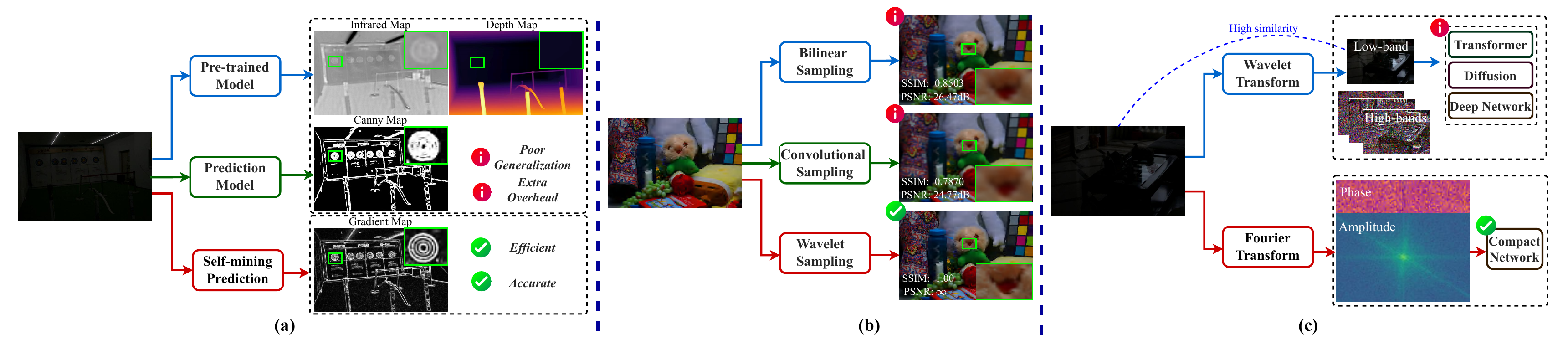}
    \vspace{-0.6cm}
    \caption{Motivation analysis.(a) Three approaches for prior generation: introducing pre-trained models to generate modal priors, adding prediction networks to predict modal priors, and self-mining priors from the image itself; (b) SSIM and PSNR computed between images after three rounds of down-and-up sampling and the original image, showing that wavelet sampling achieves lossless sampling; (c) Low-frequency information from wavelet transform requires sophisticated model processing, while amplitude and phase in the Fourier domain can be efficiently processed through compact structures.}
    \vspace{-0.3cm}
    \label{fig:motivation}
\end{figure*}

Recently, Fourier-based methods~\cite{four2,UHDFourICLR2023,zhang2024dmfourllie, feijoo2025darkir} have gained significant attention due to their efficiency in image restoration tasks. FourLLIE~\cite{four1} demonstrated that Fourier domain processing can handle global information as effectively as transformer while significantly reducing parameter overhead by replacing transformer layers in SNR-Aware~\cite{lowlight8}. However, the global nature of the Fourier transform (where spatial locality is not preserved) means that downsampling critically damages high-frequency details essential for restoration. This compels Fourier-based methods to maintain full spatial resolution, inevitably forcing a sacrifice in channel dimensions to contain model complexity, thereby restricting the expressive capacity of the model's feature space. Wavelet transform~\cite{zou2024wave,jiang2023low,zhang2025cwnet} represents another common frequency domain transformation in LLIE, capable of effectively separating content from noise. However, the low-frequency subbands retain tightly coupled spatial and frequency information, particularly containing both the dominant image content and low-frequency distortions/noise. This residual complexity, akin to a degraded version of the original image, necessitates sophisticated processing modules to effectively decouple content from corruption, offsetting the efficiency gains from wavelet decomposition.

Based on the limitations, we present our motivations:

\textit{\textbf{First}}, as shown in Fig.~\ref{fig:motivation}(a), there are two primary approaches for introducing prior clues in dark image restoration. The first approach utilizes pre-trained models to generate modal priors, where DMFourLLIE~\cite{zhang2024dmfourllie} leverages pre-trained models to generate infrared and depth map modal priors for enhanced guidance of phase component recovery, but neglects the generalization error issues introduced by pre-trained models. The second approach introduces prediction modules to generate priors, where SMG-LLE~\cite{xu2023low} designs a dedicated module to generate structural priors, but the predicted results are imprecise and introduce substantial additional overhead. This raises the question: \textit{\textbf{How can we efficiently and accurately mine high-quality endogenous priors?}}

\textit{\textbf{Second}}, as shown in Fig.~\ref{fig:motivation}(b), compared to conventional sampling, wavelet sampling possesses the advantage of lossless sampling. However, as shown in Fig.~\ref{fig:motivation}(c), the low-frequency subbands of wavelets retain tightly coupled spatial and frequency information that closely resembles the original image~\cite{jiang2023low}, still necessitating complex network architectures. In contrast, the Fourier domain can effectively handle these low-frequency components through compact structures, such as simple convolutional activations~\cite{four1}. Therefore, the question arises: \textit{\textbf{How can we combine Fourier and wavelet transforms to leverage their respective advantages?}}

Motivated by the aforementioned questions, we propose an efficient \textbf{S}elf-Mining \textbf{P}rior-Guided \textbf{J}oint \textbf{F}requency Enhancement Network (SPJFNet). First, we utilize lossless wavelet downsampling to reduce images to a lower-resolution space, thereby reducing computational overhead. In this lower-resolution space, we develop a Self-Mining Guidance Module (SMGM), which consists of a lightweight gamma enhancement network that mines the image's intrinsic gradient information to generate high-quality endogenous priors. Subsequently, we introduce a Dual Frequency Guidance Framework (DFGF) that strategically exploits the complementary strengths of wavelet and Fourier domains through a dual-branch architecture. The key innovation lies in our frequency-adaptive processing strategy: wavelet low-frequency components are transformed into the Fourier domain where compact structures can efficiently handle global illumination recovery through precise amplitude and phase reconstruction guided by self-mined priors. Meanwhile, wavelet high-frequency components remain in their native domain, where they are adaptively enhanced through a gradient-prior-guided gating mechanism that preserves fine-grained textural details. This dual-domain approach maximizes the inherent advantages of each transform while mitigating their respective limitations, achieving superior reconstruction efficiency and quality.

\begin{figure*}[t]
    \centering
    \includegraphics[width=1\linewidth]{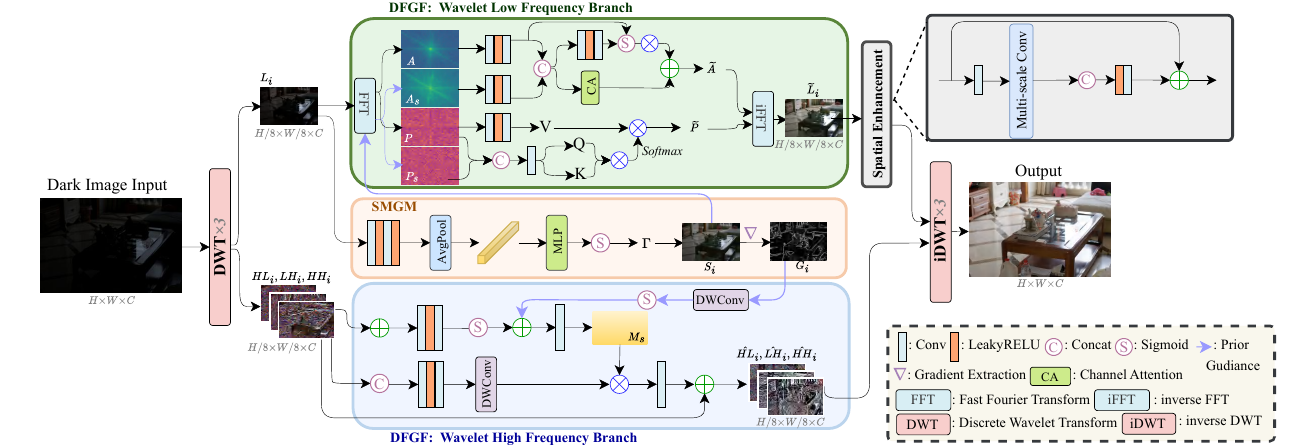}
    \vspace{-0.5cm}
    \caption{Overall framework of SPJFNet. SPJFNet employs 3-level DWT decomposition and consists of three main modules: SMGM, DFGF low-frequency branch, and DFGF high-frequency branch.}
    \vspace{-0.3cm}
    \label{fig:network}
\end{figure*}

Overall, our contributions are summarized as follows:
\vspace{-0.1cm}
\begin{itemize}
\item We introduce SMGM, which extracts endogenous priors directly from dark images, eliminating external dependencies. Our DFGF enables frequency-adaptive enhancement through prior guidance, ensuring robust reconstruction without added computational burden.

\item We propose SPJFNet, a novel dual-domain architecture that leverages the strengths of wavelet and Fourier transforms: efficiently modeling low-frequency components in the Fourier domain while preserving high-frequency details in the wavelet domain, thus achieving superior restoration with reduced complexity.

\item Extensive experiments validate our state-of-the-art performance across multiple benchmarks, demonstrating significantly reduced model complexity, as shown in Fig.~\ref{fig:ssim_computation_overhead}. Our plug-and-play framework seamlessly integrates into existing frequency-domain methods, delivering both performance gains and computational efficiency.
\end{itemize}
\section{Related Work}
\label{sec:relatedwork}

\noindent \textbf{Dark Image Restoration.}
Dark image restoration methods including histogram equalization~\cite{pizer1990contrast} and gamma correction~\cite{rahman2016adaptive}. While traditional methods are computationally efficient, they struggle with limited dynamic range modeling and significant noise amplification under extreme conditions. Recent deep learning methods~\cite{retinexformer,wu2023learning}, particularly Mamba-based techniques~\cite{bai2024retinexmamba, weng2025mamballie, guo2025mambair}, have shown promising results in Low Light Image Enhancement (LLIE). A key limitation in these approaches is the reliance on prior guidance strategies. For example, SMG-LLE~\cite{xu2023low} uses modules to predict structural priors, while SKF~\cite{wu2023learning} and DMFourLLIE~\cite{zhang2024dmfourllie} depend on pre-trained models for semantic or infrared modal priors. However, these methods face significant challenges due to imprecise predicted priors and substantial computational overhead.

\noindent \textbf{Frequency-Domain Dark Image Restoration.}
Recently, frequency-domain methods~\cite{four1, feijoo2025darkir} have become prominent for their ability to separate enhancement from noise suppression. FECNet~\cite{four2} and UHDFour~\cite{UHDFourICLR2023} utilize amplitude-phase decomposition for brightness and structural enhancement, while DMFourLLIE~\cite{zhang2024dmfourllie} employs pre-trained infrared priors for phase correction. Additionally, wavelet-based methods like Wave-Mamba~\cite{zou2024wave} and DiffLL~\cite{jiang2023low} leverage lossless downsampling and noise-content separation for texture preservation, but they require complex modules to effectively manage intricate wavelet components. A fundamental limitation of existing frequency-domain approaches is their single-domain processing: Fourier methods excel in global illumination modeling but struggle with fine details, while wavelet methods maintain local textures but necessitate complex architectures for low-frequency enhancement.

\section{Method}

\subsection{Overall Framework}
The overall framework of SPJFNet is illustrated in Fig.\ref{fig:network}. Given an input image \( I \in \mathbb{R}^{H \times W \times C} \), we use the 2D Discrete Wavelet Transform (2D-DWT) to decompose it into hierarchical frequency sub-bands:  
\begin{equation}  
\{L_i, HL_i, LH_i, HH_i\} = \text{2D-DWT}^i(I),  
\end{equation}  
where \( L_i, HL_i, LH_i, HH_i \in \mathbb{R}^{\frac{H}{2^i} \times \frac{W}{2^i} \times C} \) represent the low-frequency and high-frequency components at the \( i \)-th decomposition level. $\text{2D-DWT}$ enables efficient processing in a lower-dimensional space while preserving comprehensive frequency information.
First, the Self-Mining Guidance Module (SMGM) processes the low-frequency component through a lightweight gamma enhancement network to extract endogenous structure prior \( S_i \) and gradient priors \( G_i \), serving as guidance for both frequency branches. In the Dual Frequency Guidance Framework (DFGF), the low-frequency component \( L_i \) is transformed into the Fourier domain for global illumination recovery using amplitude and phase enhancement guided by structure prior \( S_i \). 
The enhanced components are spatially refined after inverse Fourier transform.
Meanwhile, the high-frequency components \( \{HL_i, LH_i, HH_i\} \) are processed in their native wavelet domain and enhanced through gradient-prior-guided gating mechanisms to preserve fine details. The enhanced frequency components are progressively reconstructed using multi-scale inverse wavelet transforms, benefiting from dual-domain guidance to yield a restored output image with superior quality and computational efficiency.

\begin{table*}[t]  
\centering  
\renewcommand{\arraystretch}{1.1} 
\setlength{\tabcolsep}{2.5pt} 
\resizebox{1.0\textwidth}{!}{ 
\begin{tabular}{l|l|ccc|ccc|ccc|c|c}  
\toprule  
\multirow{2}{*}{Methods} & \multirow{2}{*}{Venue} & \multicolumn{3}{c|}{LOL-v1} & \multicolumn{3}{c|}{LSRW-Huawei} & \multicolumn{3}{c|}{LSRW-Nikon} & \multirow{2}{*}{\#Param} & \multirow{2}{*}{\#FLOPs} \\   
                 &  & PSNR ↑ & SSIM ↑ & LPIPS ↓ & PSNR ↑ & SSIM ↑ & LPIPS ↓ & PSNR ↑ & SSIM ↑ & LPIPS ↓ & (M) & (G) \\ \midrule  
NPE~\cite{npe}              & TIP'13 & 16.97 &0.5928 &- & 17.08 & 0.3905 & 0.2303 & 14.86& 0.3738& 0.1464 & - & - \\
LIME~\cite{lime}            & TIP'16 & 16.75 &0.4440& 0.2060 & 17.00 & 0.3816 & 0.2069 & 13.53 &0.3321 &0.1272 & - & - \\
SRIE~\cite{retinex1}        & ICCV'16 & 11.80 &0.5000 &0.1862 & 13.42 & 0.4282 & 0.2166 & 13.26 &0.3963 &0.1396 & - & - \\
Kind~\cite{kind}            & MM'19 & 20.87& 0.7995 &0.2071 & 16.58 & 0.5690 & 0.2259 &11.52 &0.3827 &0.1860 & 8.02 & 34.99 \\
MIRNet~\cite{lowlight9}     & ECCV'20 & \underline{24.14} &0.8305& 0.2502 & 19.98 & 0.6085 & 0.2154 & 17.10 & 0.5022 & 0.2170 & 31.79 & 785.1 \\
Kind++~\cite{kind++}        & IJCV'21 & 18.97 &0.8042 &0.1756 & 15.43 & 0.5695 & 0.2366 & 14.79 & 0.4749 & \underline{0.2111} & 8.27 & - \\
SNR-Aware~\cite{lowlight8}  & CVPR'22 & 23.93 &0.8460& 0.0813 & 20.67 & 0.5911 & 0.1923 & 17.54 & 0.4822 & 0.0982 & 39.12 & 26.35 \\
FourLLIE~\cite{four1}       & MM'23 & 20.99 & 0.8071 & 0.0952 & 21.11 & 0.6256 & 0.1825 & 17.82 & 0.5036 & 0.2150 & \textbf{0.12} & 4.07 \\
UHDFour~\cite{UHDFourICLR2023} & ICLR'23 & 22.89 & 0.8147 & 0.0934 & 19.39 & 0.6006 & 0.2466 & \underline{17.94} & 0.5195 & 0.2546 & 17.54 & 4.78 \\
Retinexformer~\cite{retinexformer} & ICCV'23 & 22.71 & 0.8177 & 0.0922 & 21.23 & 0.6309 & 0.1699 & 17.64 & 0.5082 & 0.2784 & 1.61 & 15.57 \\ 
Wave-Mamba~\cite{zou2024wave} & MM'24 & 22.76 & 0.8419 & 0.0791 & 21.19 & 0.6391 & 0.1818 & 17.34 & 0.5192 & 0.2294 & 1.26 & 7.22 \\ 
UHDFormer~\cite{wang2024uhdformer} & AAAI'24 & 22.88	&0.837	&0.1390 & 20.64 & 0.6244 & 0.1812 & 17.01 & 0.5186 & 0.2793 & 0.34 & \underline{3.24} \\ 
DMFourLLIE~\cite{zhang2024dmfourllie} & MM'24 & 22.98 &  0.8273  & 0.0792 & 21.47 & 0.6331 & 0.1781 & 17.04 & \underline{0.5274} & 0.2178 & 0.75 & 5.81 \\ 
Retinexmamba~\cite{bai2024retinexmamba} & ICONIP'24 & 23.15 & 0.8210 & 0.0876 & 20.88 & 0.6298 & 0.1689 & 17.59 & 0.5133 & 0.2534 & 3.59 & 34.76 \\ 
CWNet~\cite{zhang2025cwnet} & ICCV'25 & \underline{23.60} & 0.8496 & \textbf{0.0648} & \underline{21.50} & \underline{0.6397} & \underline{0.1562} & 17.38 & 0.5119 & 0.2575 & 1.23 & 11.3 \\
CIDNet~\cite{yan2025hvi} & CVPR'25 & 23.81 & \textbf{0.8574} & 0.0856 & 20.30 & 0.6054 & 0.3500 & 17.16 & 0.4975 & 0.2816 & 1.88 & 7.57 \\ 
 \midrule
SPJFNet & Ours & \textbf{24.16} & \underline{0.8531} & \underline{0.0694} & \textbf{21.71} & \textbf{0.6547} & \textbf{0.1558} & \textbf{17.98} & \textbf{0.5289} & \textbf{0.2047} & \underline{0.21} & \textbf{2.43} \\ 
\bottomrule  
\end{tabular}  }
\vspace{-0.1cm}
\caption{Quantitative comparison on LOL-v1, LSRW-Huawei and LSRW-Nikon datasets without using ground truth mean. The best results are highlighted in \textbf{bold} and the second-best results are \underline{underlined}.}  
\vspace{-0.2cm}
\label{tab:com-combined}  
\end{table*}

\subsection{Self-Mining Guidance Module (SMGM)}
As illustrated in Fig.\ref{fig:motivation}(a), our self-mining prediction approach demonstrates that gradient maps predicted through Gamma transformation of dark images exhibit clear structural details. This endogenous self-mining strategy is both precise and efficient, avoiding the error propagation and computational burden associated with handcrafted or complex prediction modules.
Specifically, SMGM operates on the low-frequency component $L_i$ to extract dual complementary priors: the enhanced structural prior $S_i$ and gradient prior $G_i$, which serve as endogenous guidance for the dual-frequency enhancement framework. $L_i$ is first processed through convolutional layers to extract feature representations, followed by global average pooling to obtain the global feature vector $\mathbf{F}_{\text{a}}$, providing channel-wise global representation. $\mathbf{F}_{\text{a}}$ is then fed into a multi-layer perceptron (MLP) followed by a Sigmoid activation function to produce the adaptive Gamma mapping $\Gamma$:

\begin{equation}  
\Gamma = \sigma(\text{MLP}(\mathbf{F}_{\text{a}})), \quad \Gamma \in \mathbb{R}^{1 \times 1 \times C},
\end{equation}

where $\sigma$ denotes the Sigmoid activation function that constrains output values to $(0, 1)$. The Gamma mapping is applied to generate the enhanced structural prior $S_i$, from which the gradient prior $G_i$ is extracted using the Sobel operator~\cite{kanopoulos1988design}:
\begin{equation}
S_i = L_i^{\Gamma}, \quad G_i = \nabla(S_i).
\end{equation}

The key insight lies in the dual-guidance mechanism: $S_i$ serves as the illumination-enhanced prior that guides the DFGF wavelet low-frequency branch for global brightness recovery, while $G_i$ provides structural gradient information that guides the DFGF wavelet high-frequency branch for detail preservation. This complementary design ensures that when the gradient prior $G_i$ achieves high accuracy through our supervised learning objective, the corresponding Gamma-enhanced structural prior $S_i$ inherently possesses reliable illumination guidance capabilities, as both priors originate from the same adaptive Gamma transformation.
To ensure the accuracy of $G_i$, the learning objective of SMGM incorporates two complementary losses: gradient fidelity loss and structural consistency loss:
\begin{equation}
{\mathcal L}_{s} = \lambda_1 \cdot \|G_i - G_{i}^{gt}\|_1 + \lambda_2 \cdot \text{CE}(\text{Edge}(S_i), \text{Edge}(S_{i}^{gt})),
\label{eq:Lg}
\end{equation}
where $\lambda_1 = 1.0$ and $\lambda_2 = 0.1$ are weighting coefficients, $\text{CE}$ represents cross-entropy loss for edge maps, and $\text{Edge}(\cdot)$ extracts edge information. ${\mathcal L}_{s}$ ensures that both the gradient prior $G_i$ and structural prior $S_i$ maintain high fidelity to ground truth, establishing a robust foundation for the subsequent dual-frequency enhancement process.

\subsection{Dual-Frequency Guidance Framework (DFGF)}

\subsubsection{Wavelet Low Frequency Branch.}
While wavelet decomposition provides efficient multi-scale downsampling, its low-frequency subband inherently retains coupled spatial-frequency information, including residual high-frequency components that impede pure frequency analysis. To leverage both the structural preservation of lossless, we strategically transform wavelet low-frequency components into the Fourier domain.

As illustrated in Fig.\ref{fig:network}, the low-frequency component $L_i$ is transformed into the Fourier domain using Fast Fourier Transform (FFT), yielding amplitude $A$ and phase $P$: $A, P = \text{FFT}(L_i)$. Simultaneously, the structural prior $S_i$ encoding precise brightness and structural information is transformed into amplitude prior $A_s$ and phase prior $P_s$: $A_s, P_s = \text{FFT}(S_i)$.

Since prior work~\cite{four1} indicates that brightness information is mainly encoded in the amplitude component, we propose a dual-branch architecture to leverage this. We first concatenate the amplitude prior $A_s$ with the original amplitude $A$ to form enriched features, $A_{concat} = \text{Concat}(A, A_s)$, which are then processed through complementary branches. The first branch modulates the original amplitude component using channel attention mechanisms to highlight salient features, while the second branch employs gating mechanisms for precise mapping of the amplitude components. This dual-path processing synergistically enhances global illumination recovery. This process can be formulated as:
\begin{equation}
\text{B}_1 = \text{CA}(A_{concat}) , \quad \text{B}_2 = \sigma(\text{Conv}(A_{concat})) \cdot A,
\end{equation}

where $\text{CA}(\cdot)$ denotes channel attention, and $\sigma$ is the Sigmoid activation function. $\text{B}_1$ applies channel attention to the concatenated features and modulates the original amplitude, enabling adaptive amplitude adjustment based on both original and prior information. $\text{B}_2$ processes the amplitude prior through convolution and Sigmoid activation to extract refined prior features with values constrained to $(0,1)$ for stable enhancement. The enhanced amplitude is obtained through element-wise addition: $\tilde{A} = \text{B}_1 + \text{B}_2$, where the addition operation preserves the original amplitude characteristics while incorporating prior-guided enhancements.

Phase components contain crucial high-frequency structural details that require careful preservation during enhancement. Unlike amplitude components that primarily encode global illumination, phase components capture local structural information that benefits from attention-based refinement. The phase prior $P_s$ provides structural guidance for enhancing the original phase $P$ through a self-attention framework:
\begin{equation}
\begin{aligned}
Q, K &= \text{Conv}(\text{Concat}(P_s, P)), \quad V = \text{Conv}(P), \\
\tilde{P} &= \text{Softmax}(Q \cdot K^\top) \cdot V,
\end{aligned}
\end{equation}
where the query $Q$ and key $K$ are generated from the concatenated phase and phase prior to capture correlations between structural patterns, while the value $V$ is derived from the original phase to preserve authentic structural information. The self-attention mechanism selectively enhances phase components based on structural similarity, ensuring coherent phase reconstruction.
The enhanced amplitude $\tilde{A}$ and phase $\tilde{P}$ are then merged and transformed back to the spatial domain via inverse FFT.


Following the two-stage processing paradigm of current Fourier-based methods, we employ a Spatial Enhancement Module to further refine local spatial details. The core component consists of two complementary convolution modules with different receptive field characteristics: dilated convolution~\cite{yu2015multi} with dilation rate 4 for expanded receptive field coverage, and wavelet transform convolution (WTConv)~\cite{finder2025wavelet} with 3 decomposition levels for multi-scale feature representation. The dilated convolution captures long-range spatial dependencies through enlarged receptive fields, while the WTConv leverages wavelet decomposition to extract multi-resolution features across different frequency bands. This dual-branch design achieves comprehensive spatial detail enhancement while maintaining computational efficiency, effectively recovering fine-grained spatial information.

\begin{table}[t]  
\centering  
\renewcommand{\arraystretch}{1.2} 
\setlength{\tabcolsep}{4pt} 
\resizebox{1.0\columnwidth}{!}{ 
\begin{tabular}{l|ccc}  
\toprule  
Methods & \multicolumn{3}{c}{UHD-LL} \\ & PSNR ↑ & SSIM ↑ & LPIPS ↓  \\ \midrule 
LLFormer~\cite{jie2023llformer}       & 22.79     & 0.8530    & 0.2640          \\ 
FourLLIE~\cite{four1}         & 23.96           & 0.8826     & 0.1261          \\    
UHDFour~\cite{UHDFourICLR2023}    & \textbf{26.22} &0.900 &0.239        \\  
WaveMamba~\cite{zou2024wave}    & 25.82          & 0.9221      & 0.1092         \\ 
UHDFormer~\cite{wang2024uhdformer}  & 23.98&0.9017&0.0934\\
RetinexMamba~\cite{bai2024retinexmamba}  & 24.39&0.9078&0.1372\\
DMFourLLIE~\cite{zhang2024dmfourllie} & 24.76&\underline{0.9174}&0.1145\\
CWNet~\cite{zhang2025cwnet} & 24.27& 0.9024 &\underline{0.057}\\
\midrule 
Ours      & \underline{25.98}  & \textbf{0.9223} & \textbf{0.052} \\  
\bottomrule  
\end{tabular}}  
\caption{Quantitative comparison on the UHD-LL dataset.}  
\label{tab:com-uhd}  
\end{table}  

\begin{table}[t]  
\centering  
\renewcommand{\arraystretch}{1.1} 
\setlength{\tabcolsep}{8pt} 
\resizebox{1\columnwidth}{!}{%
\begin{tabular}{l|ccccc}  
\toprule 
Methods & DICM & LIME & MEF & NPE & VV \\   
\midrule  
Kind~\cite{kind} & \textbf{3.61} & 4.77 & 4.82 & 4.18 & 3.84 \\   
MIRNet~\cite{lowlight9} & 4.04 & 6.45 & 5.50 & 5.24 & 4.74 \\   
SGM~\cite{lol} & 4.73 & 5.45 & 5.75 & 5.21 & 4.88 \\   
FECNet~\cite{four2} & 4.14 & 6.04 & 4.71 & 4.50 & 3.75 \\   
HDMNet~\cite{liang2022learning} & 4.77 & 6.40 & 5.99 & 5.11 & 4.46 \\   
Bread~\cite{guo2023low} & 4.18 & 4.72 & 5.37 & 4.16 & \textbf{3.30} \\   
Retinexformer~\cite{retinexformer} & 4.01 & \underline{3.44} & 3.73 & 3.89 & 3.71 \\   
UHDFormer~\cite{wang2024uhdformer} & 4.42 & 4.35 & 4.74 & 4.40 & 4.28 \\
Wave-Mamba~\cite{zou2024wave} & 4.56 & 4.45 & 4.76 & 4.54 & 4.71 \\    
CWNet~\cite{zhang2025cwnet} &3.92 & 3.58 & \underline{3.66} & \underline{3.61} & 3.74 \\    \midrule  
Ours & \underline{3.85} & \textbf{3.38} & \textbf{3.69} & \textbf{3.52} & \underline{3.62} \\ 
\bottomrule  
\end{tabular}}  
\caption{NIQE scores on DICM, LIME, MEF, NPE, and VV datasets. All methods are pre-trained on LSRW-Huawei.}  
\label{tab:comsmall}  
\end{table}

\begin{figure*}[t]
    \centering
    \includegraphics[width=1\linewidth]{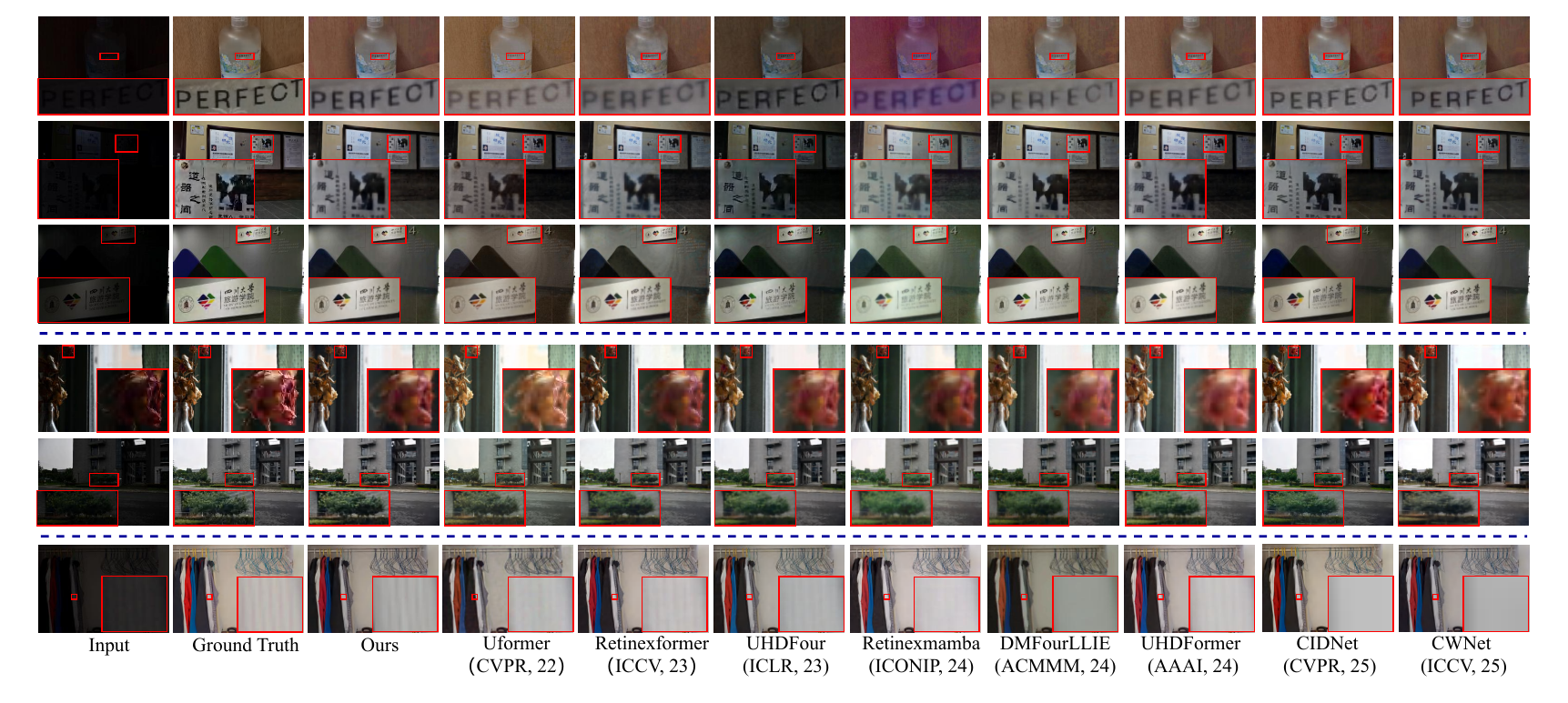}
    \vspace{-0.6cm}
    \caption{Visual comparison with state-of-the-art methods across three datasets. From top to bottom: three groups of comparisons are conducted on LSRW-Huawei, LSRW-Nikon, and LOL-v1 datasets, respectively.}
    \vspace{-0.4cm}
    \label{fig:huawei}
\end{figure*}

\begin{figure}[t]
    \centering
    \includegraphics[width=1\linewidth]{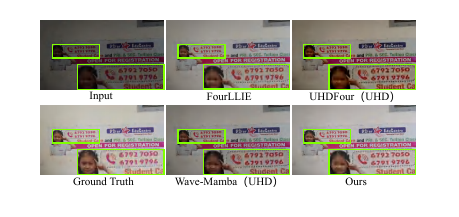}
    \caption{Visual comparison on UHD-LL dataset. (UHD) is a designed for UHD images.}
    \label{fig:UHD}
\end{figure}

\begin{figure}[t]
    \centering
    \includegraphics[width=1\linewidth]{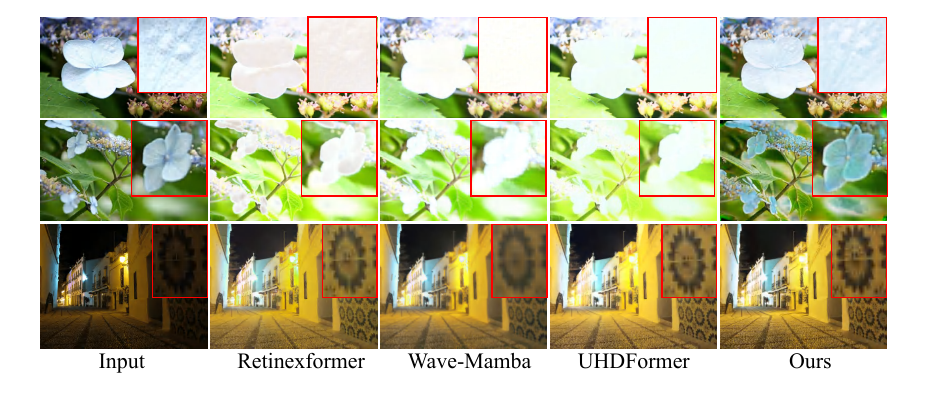}
    \caption{Visual comparison on no-reference  datasets.}
    \label{fig:small}
\end{figure}

\subsubsection{Wavelet High Frequency Branch.}
High-frequency components primarily encode image edges, textures, and fine-grained details that require precise local enhancement to preserve structural integrity while avoiding noise amplification. Our gradient-guided enhancement strategy leverages the extracted gradient prior $G_i$ to provide structural guidance for adaptive high-frequency enhancement.

As illustrated in Fig.\ref{fig:network}, we first fuse the gradient prior $G_i$ with the high-frequency components through element-wise addition to create structurally-informed features: $\mathbf{F}_{hf} = (HL_i + LH_i + HH_i) + G_i$. Simultaneously, $G_i$ is processed through depthwise separable convolution to extract efficient spatial features: $\mathbf{F}_{dw} = \text{DWConv}(G_i)$.

To generate precise enhancement weights, the extracted gradient features are transformed into a spatial gating map through 1×1 convolution and Sigmoid activation:
\begin{equation}
M_s = \sigma(\text{Conv}_{1\times1}(\mathbf{F}_{dw})),
\end{equation}
where $M_s \in (0,1)$ serves as a gradient-guided gating map that indicates enhancement confidence at each spatial location. Regions with strong gradient responses receive higher enhancement weights, while smooth regions are preserved with minimal modification.

The high-frequency components $\{HL_i, LH_i, HH_i\}$ are concatenated and processed through depthwise separable convolution to extract spatial features: $\mathbf{F}_{spatial} = \text{DWConv}(\text{Concat}(HL_i, LH_i, HH_i))$. The gradient-guided gating mechanism is applied for selective enhancement: $\mathbf{F}_{enhanced} = \mathbf{F}_{spatial} \odot M_s + \mathbf{F}_{spatial}$, where the residual connection preserves original information while enabling adaptive enhancement.
Finally, the enhanced features are processed through convolution and residual blocks, then decomposed back into refined high-frequency components $\{\hat{HL}_i, \hat{LH}_i, \hat{HH}_i\}$ through learned projection layers, ensuring structurally-consistent detail enhancement for subsequent wavelet reconstruction.

\section{Experiments}
\subsection{Datasets and Experimental Details}  
\noindent \textbf{Datasets.} We train and evaluate our method on four widely recognized dark image datasets: LOL~\cite{lol}, LSRW-Huawei~\cite{lsrw}, LSRW-Nikon~\cite{lsrw}, and UHD-LL~\cite{UHDFourICLR2023}. The LOL dataset includes 485 training pairs and 15 testing pairs. LSRW-Huawei contains 3,150 training pairs and 20 testing pairs, while LSRW-Nikon provides 2,450 training pairs and 30 testing pairs. UHD-LL~\cite{UHDFourICLR2023} is a high-resolution dataset with 2,000 training pairs and 115 testing pairs. Additionally, we evaluate on three unpaired datasets: NPE~\cite{npe}, DICM~\cite{dicm}, LIME~\cite{lime}, MEF~\cite{mefl} and VV~\cite{vonikakis2018evaluation}.

\noindent \textbf{Implementation Details.} Our method is implemented in PyTorch and trained end-to-end. Images are randomly cropped to $256 \times 256$ and augmented with random flipping. We use the ADAM optimizer with an initial learning rate of $4.0\times10^{-4}$ and a multi-step scheduler. Training is conducted for $1.5\times10^{5}$ iterations with a batch size of 8 on two NVIDIA 4090 GPUs. Loss function utilizes ${\mathcal L}_{1}$ loss for training.

\noindent \textbf{Evaluation.}
Our network performance is primarily evaluated using peak signal-to-noise ratio (PSNR~\cite{hore2010image}), structural similarity index (SSIM~\cite{wang2004image}) and LPIPS~\cite{zhang2018unreasonable}.
We conduct comprehensive comparisons with 16 state-of-the-art dark image restoration methods, including traditional approaches (NPE~\cite{npe}, LIME~\cite{lime}, SRIE~\cite{retinex1}), deep learning-based methods (Kind~\cite{kind}, MIRNet~\cite{lowlight9}, Kind++~\cite{kind++}, SNR-Aware~\cite{lowlight8}, FourLLIE~\cite{four1}, UHDFour~\cite{UHDFourICLR2023},Retinexformer~\cite{retinexformer}, Wave-Mamba~\cite{zou2024wave}, UHDFormer~\cite{wang2024uhdformer}, DMFourLLIE~\cite{zhang2024dmfourllie}, Retinexmamba~\cite{bai2024retinexmamba}, CWNet~\cite{zhang2025cwnet} and CIDNet~\cite{yan2025hvi}).

\begin{table}[t]  
\centering  
\renewcommand{\arraystretch}{1.1}
\setlength{\tabcolsep}{4pt}
\resizebox{1\columnwidth}{!}{%
\begin{tabular}{c|ccccc|cc}  
\toprule
Method & $\text{DWT}_{i}$ & ${\mathcal L}_{s}$ & SMGM & D-L & D-H & PSNR↑ & SSIM↑ \\
\midrule
A & 2 & \checkmark & \checkmark & \checkmark & \checkmark & 21.13 & 0.64 \\
B & 4 & \checkmark & \checkmark & \checkmark & \checkmark & \textbf{21.73} & \underline{0.6535} \\
\midrule
C & 3 & $\times$ & \checkmark & \checkmark & \checkmark & $Nan$ & $Nan$ \\
\midrule
D & 3 & \checkmark & $\times$ & \checkmark & \checkmark & 20.98 & 0.6384 \\
E & 3 & \checkmark & \checkmark & $\times$ & \checkmark & 20.84 & 0.6371 \\
F & 3 & \checkmark & \checkmark & \checkmark & $\times$ & 21.30 & 0.6523 \\
G & 3 & \checkmark & $\times$ & \checkmark & $\times$ & 20.65 & 0.6346 \\
H & 3 & \checkmark & \checkmark & $\times$ & $\times$ & 20.47 & 0.6331 \\
I & 3 & \checkmark & $\times$ & $\times$& \checkmark & 20.82 & 0.6405 \\
\midrule
J & 3 & \checkmark & \checkmark & \checkmark & \checkmark & \underline{21.71} & \textbf{0.6547} \\
\bottomrule  
\end{tabular}}  
\caption{Ablation study on core architectural components.  \checkmark indicates the component is used, $\times$ indicates it is not used.}  
\label{tab:ablation1}  
\end{table}

\begin{table}[t]
\centering  
\renewcommand{\arraystretch}{1.1}
\setlength{\tabcolsep}{4pt}
\resizebox{1\columnwidth}{!}{%
\begin{tabular}{ccc|ccc|cc|cc|cc}  
\toprule
\multicolumn{3}{c|}{D-L} & \multicolumn{3}{c|}{D-H} & \multicolumn{2}{c|}{M-s} & \multicolumn{2}{c|}{${\mathcal L}_{s}$} & \multicolumn{2}{c}{Metrics} \\
\midrule
$Amp$ & $Pha$ & $Spa$ & $M_{s}$ & $F_{hf}$ & $F_{s}$ & WTC & DC & $\lambda_1$ & $\lambda_2$ & PSNR↑ & SSIM↑ \\
\midrule
$\times$ & \checkmark & \checkmark & \checkmark & \checkmark & \checkmark & \checkmark & \checkmark & 1.0 & 0.1 &  21.23 & 0.6497 \\
\checkmark & $\times$ & \checkmark & \checkmark & \checkmark & \checkmark & \checkmark & \checkmark & 1.0 & 0.1 & 21.58 & 0.6532 \\
\checkmark & \checkmark & $\times$ & \checkmark & \checkmark & \checkmark & \checkmark & \checkmark & 1.0 & 0.1 &  21.28 & 0.6488 \\
\midrule
\checkmark & \checkmark & \checkmark & $\times$ & \checkmark & \checkmark & \checkmark & \checkmark & 1.0 & 0.1 & 21.42 & 0.6527 \\
\checkmark & \checkmark & \checkmark & \checkmark & $\times$ & \checkmark & \checkmark & \checkmark & 1.0 & 0.1 &  21.54 & 0.6543 \\
\checkmark & \checkmark & \checkmark & \checkmark & \checkmark & $\times$ & \checkmark & \checkmark & 1.0 & 0.1 &  21.42 & 0.6527 \\
\midrule
\checkmark & \checkmark & \checkmark & \checkmark & \checkmark & \checkmark & $\times$ & \checkmark & 1.0 & 0.1 &  21.30 & 0.6495 \\
\checkmark & \checkmark & \checkmark & \checkmark & \checkmark & \checkmark & \checkmark & $\times$ & 1.0 & 0.1 &  21.41 & 0.6502 \\
\midrule
\checkmark & \checkmark & \checkmark & \checkmark & \checkmark & \checkmark & \checkmark & \checkmark & 0.8 & 0.1 &  21.69 & 0.6539 \\
\checkmark & \checkmark & \checkmark & \checkmark & \checkmark & \checkmark & \checkmark & \checkmark & 1.0 & 0.2 &  21.64 & 0.6528 \\ \midrule
\checkmark & \checkmark & \checkmark & \checkmark & \checkmark & \checkmark & \checkmark & \checkmark & 1.0 & 0.1 &  \textbf{21.71} & \textbf{0.6547} \\
\bottomrule  
\end{tabular}}  
\caption{Ablation study on fine-grained components.}  
\label{tab:ablation2}  
\end{table}

\begin{table}[t]  
\centering  
\renewcommand{\arraystretch}{1.2} 
\setlength{\tabcolsep}{4pt} 
\resizebox{\columnwidth}{!}{ 
\begin{tabular}{l|ccc}  
\toprule  
\textbf{Methods} & \textbf{PSNR ↑} & \textbf{SSIM ↑} & \textbf{LPIPS ↓}   \\ \midrule 
FourLLIE (Baseline)  & 21.11    & 0.6256   & 0.1825        \\  
FourLLIE + SMGM      & \underline{21.37}   & \underline{0.6401}   & \underline{0.1688}        \\   
FourLLIE + SMGM + DFGF      & \textbf{21.49}   & \textbf{0.6448}   & \textbf{0.1632}       \\
\midrule  
DMFourLLIE (Baseline)  & 21.47    & 0.6331    & 0.1781        \\  
DMFourLLIE + SMGM    & \underline{21.52}     & \underline{0.6429}  & \underline{0.1597}          \\  
DMFourLLIE + SMGM + DFGF     & \textbf{21.70}     & \textbf{0.6457}  & \textbf{0.1601}          \\  
\bottomrule  
\end{tabular}}  
\caption{Plug-and-play validation on Fourier-based methods.}  
\label{tab:plug_test}  
\end{table}  

\subsection{Comparisons with State-of-the-Art Methods}
\textbf{Quantitative comparison on LOL-V1, LSRW-Huawei and LSRW-Nikon datasets.} The quantitative results are presented in Tab.~\ref{tab:com-combined}. On the LOL-V1 dataset, our method achieves the best PSNR while maintaining competitive SSIM (0.8531) and LPIPS (0.0694) scores. For the LSRW-Huawei dataset, our method obtains the highest PSNR (21.71 dB) and SSIM (0.6547), with the best LPIPS score (0.1558). Similarly, on the LSRW-Nikon dataset, our method achieves the best performance across all metrics. \textbf{Notably}, our method excels in computational efficiency while maintaining superior performance. With only 0.21M parameters and 2.43G FLOPs, our approach is significantly more lightweight compared to competing methods.

\noindent \textbf{Quantitative Comparison on UHD-LL Dataset.} The quantitative results on the ultra-high-definition low-light dataset are presented in Tab.~\ref{tab:com-uhd}. Our method demonstrates exceptional performance across multiple evaluation metrics when compared with 8 state-of-the-art methods.

\noindent \textbf{Quantitative Comparison on No-reference Datasets.} As shown in Tab.\ref{tab:comsmall}, our method obtains the best NIQE scores on three out of five datasets (LIME, MEF, NPE) and ranks second on the remaining two datasets (DICM, VV), consistently outperforming existing state-of-the-art methods.

\noindent \textbf{Visual Comparisons.} To demonstrate the effectiveness of our approach, we present visual comparisons with state-of-the-art methods across three challenging datasets in Fig.~\ref{fig:huawei}. The results clearly illustrate the superior enhancement quality achieved by our method in various challenging scenarios. Fig.~\ref{fig:UHD} shows the visual comparisons on the UHD-LL dataset. Compared to other methods, our approach produces results that are closer to the ground truth in terms of brightness, image quality, and detail preservation. Fig.~\ref{fig:small} shows visual comparisons on the unpaired DICM and LIME datasets, which demonstrates our method's superior performance in detail preservation and exposure control. 

\subsection{Ablation Study}
\noindent \textbf{Core Component Analysis.} Tab.~\ref{tab:ablation1} presents the ablation study on core architectural components. We investigate the optimal wavelet decomposition levels (rows A-B) and validate the necessity of loss ${\mathcal L}_{s}$ (row C), whose removal causes training instability (NaN values). Individual module analysis (rows D-I) shows SMGM contributes 0.73 dB PSNR improvement, while DFGF low-frequency and high-frequency (D-L, D-H) branches provide 0.87 dB and 0.41 dB improvements respectively. The combination analysis (rows G-I) confirms their synergistic effect.

\noindent \textbf{Fine-grained Component Analysis.} Tab.~\ref{tab:ablation2} analyzes sub-components within each module. In the DFGF low-frequency branch (D-L), amplitude guidance contributes most significantly (0.48 dB), followed by spatial enhancement (0.43 dB) and phase guidance (0.13 dB). For the high-frequency branch (D-H), gradient map guidance ($M_s$) provides 0.29 dB improvement, while feature processing components $F_{hf}$ and $F_s$ contribute 0.17 dB and 0.29 dB respectively. In the multi-scale module, WTConv and dilated convolution contribute 0.41 dB and 0.30 dB respectively, demonstrating their complementary roles. Additionally, we conducted an ablation study to verify the weight of 
${\mathcal L}_{s}$.

\noindent \textbf{Plug-and-Play Validation.} To further demonstrate the generalizability and effectiveness of our proposed modules, we conduct plug-and-play experiments by integrating our components into existing Fourier-based low-light enhancement methods, specifically FourLLIE and DMFourLLIE. As shown in Tab.~\ref{tab:plug_test}, both methods achieve consistent performance improvements with our modules.

\section{Limitation and Conclusion}
\noindent \textbf{Limitation.}
While our method demonstrates superior performance in exposure control and detail preservation, there are certain limitations. As illustrated in the second row of Fig.~\ref{fig:small}, our method occasionally exhibits color distortion where the restored image shows slight color shifts compared to natural appearance.
This limitation stems from our frequency-domain processing strategy, which emphasizes amplitude and phase information while color consistency constraints could be strengthened. The gradient-guided enhancement, though effective for spatial detail recovery, may prioritize structural information over color accuracy in challenging scenarios, suggesting room for improvement in color fidelity preservation, particularly in complex lighting conditions.

\noindent \textbf{Conclusion.}
In this paper, we present SPJFNet, an efficient self-mining prior-guided joint frequency enhancement network. By introducing the SMGM to eliminate dependence on external priors and the DFGF to decouple frequency processing, our method achieves significant efficiency improvements while surpassing state-of-the-art performance. In the future, we will focus our research efforts on how to efficiently combine frequency domain processing with ensuring color consistency in dark image restoration.

\bibliography{aaai2026}


\setcounter{section}{0}
\renewcommand{\thesection}{\Alph{section}}  
\clearpage
\setcounter{page}{1}

\twocolumn[ 
    \begin{center}
        \textbf{\Large Supplementary Material}
    \end{center}
    \vspace{1em}  
]
\setcounter{figure}{0}
\setcounter{table}{0}

\section{Visualization of Intermediate Processes and Comparative Analysis}

\subsection{Visualization of Intermediate Processes}

\begin{figure*}[t]
    \centering
    \includegraphics[width=1\linewidth]{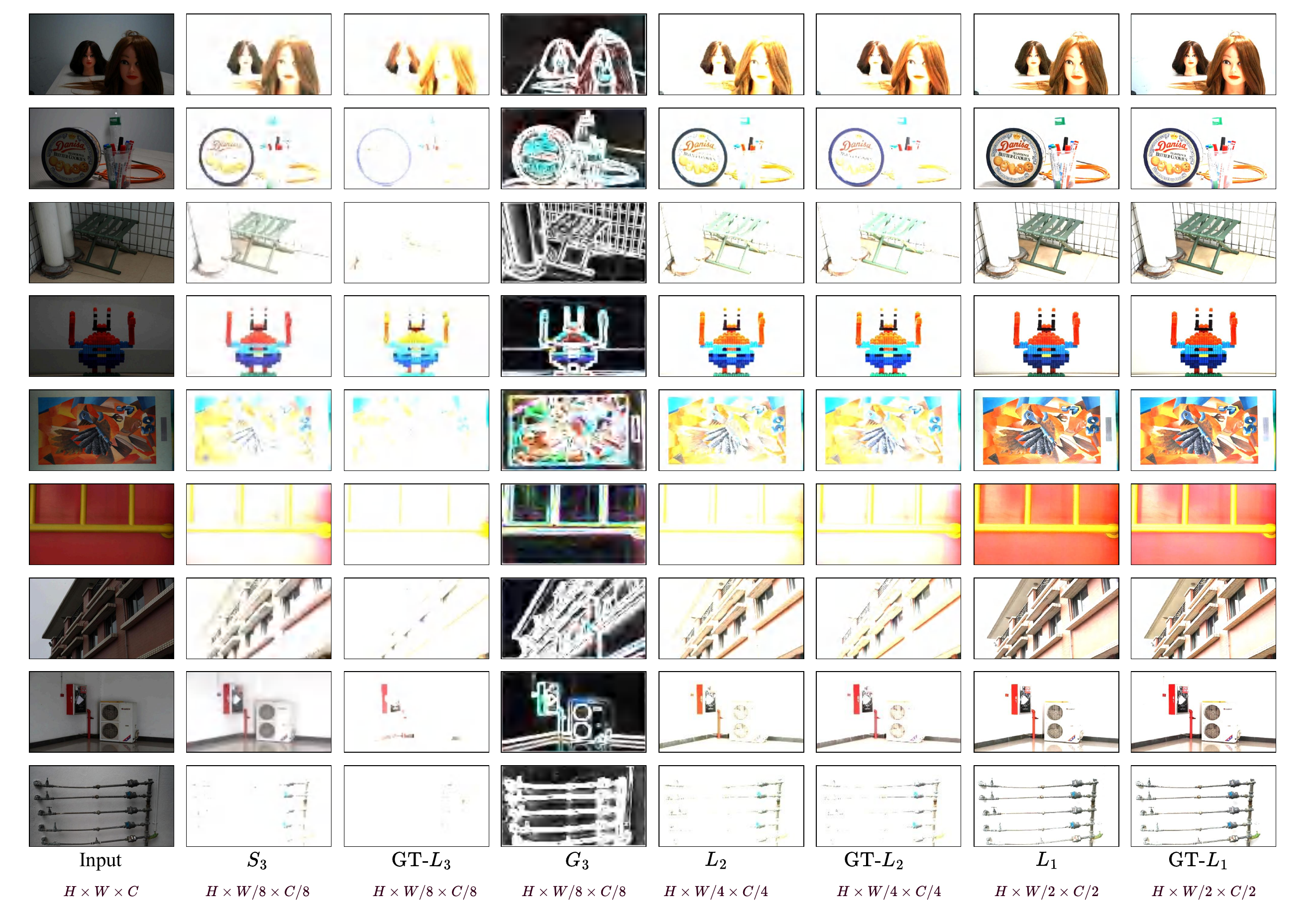}
    \caption{Visualization of endogenous priors and intermediate results on the UHD-LL dataset. The second and fourth columns highlight the self-mined endogenous prior \(S_3\) and its gradient prior \(G_3\), showcasing superior content representation and edge detail recovery compared to the GT-$L_{3}$ baseline.}
    \label{fig:media_show_uhd}
\end{figure*}

\begin{figure*}[t]
    \centering
    \includegraphics[width=1\linewidth]{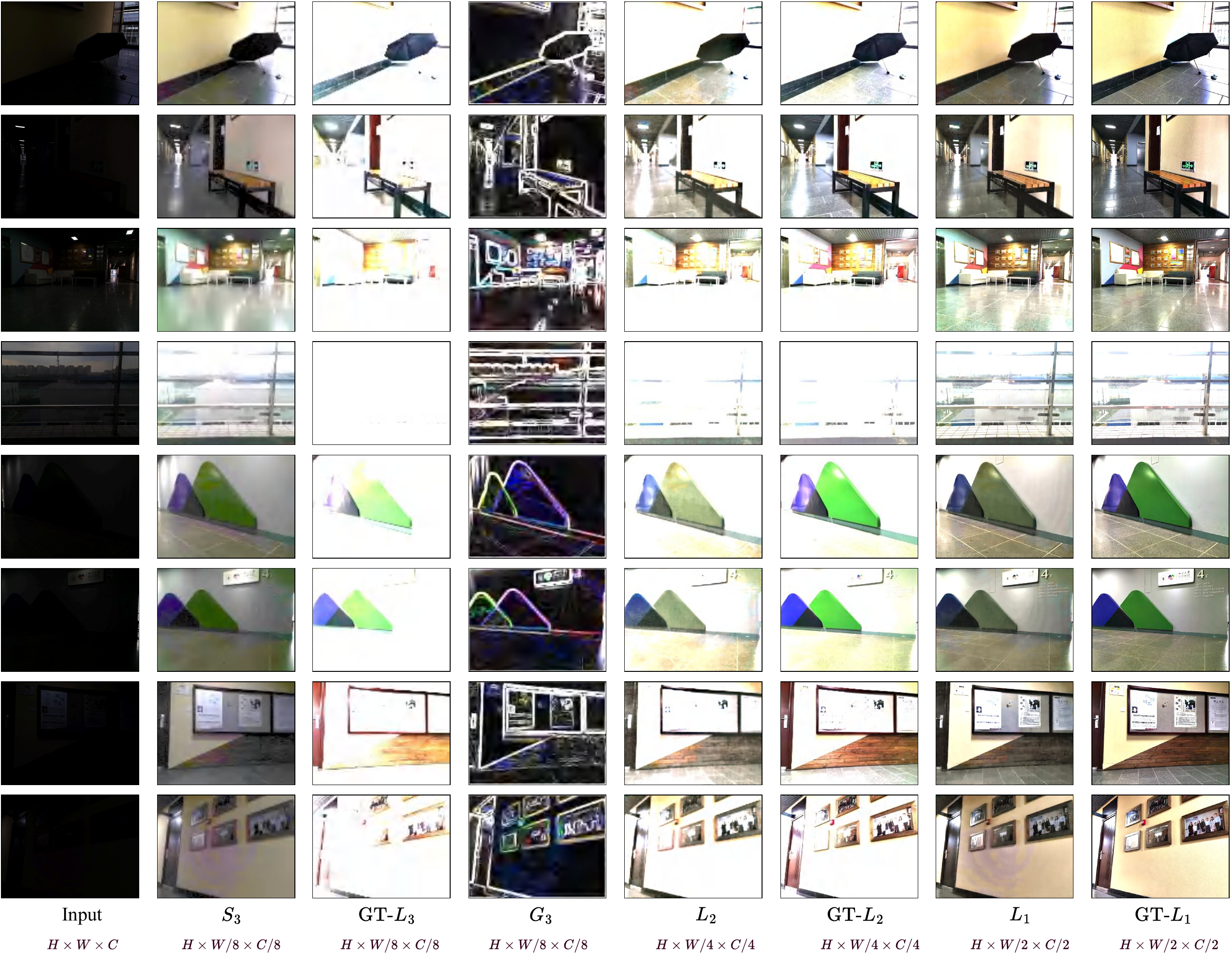}
    \caption{Visualization of endogenous priors and intermediate results on the LSRW-Huawei dataset.}
    \label{fig:media_show_huawei}
\end{figure*}

In Fig.\ref{fig:media_show_uhd} and Fig.\ref{fig:media_show_huawei}, we present a comparative visualization of endogenous priors and intermediate results. The first row includes the input images, while the second and fourth columns showcase the endogenous prior \(S_3\) and its gradient prior \(G_3\), extracted via SMGM after three-wavelet downsampling at the resolution \(H/8 \times W/8 \times C\). In contrast, $\text{GT-}L_{3}$ is derived from ground truth (GT) wavelet downsampling. Notably, \(S_3\) captures a significantly richer content prior compared to $\text{GT-}L_{3}$, and the gradient prior \(G_3\) exhibits precise high-frequency contour details, which allows for more accurate edge representation and texture recovery. Subsequent visual comparisons following the second and first layer wavelet sampling reveal that our method closely aligns with ground truth (GT) images in terms of brightness, color fidelity, and texture details, showcasing its effectiveness in dark conditions. The ability to visualize these components allows for a better understanding of the restoration dynamics.

\subsection{Additional Visual Comparisons}

\begin{figure*}[t]
    \centering
    \includegraphics[width=1\linewidth]{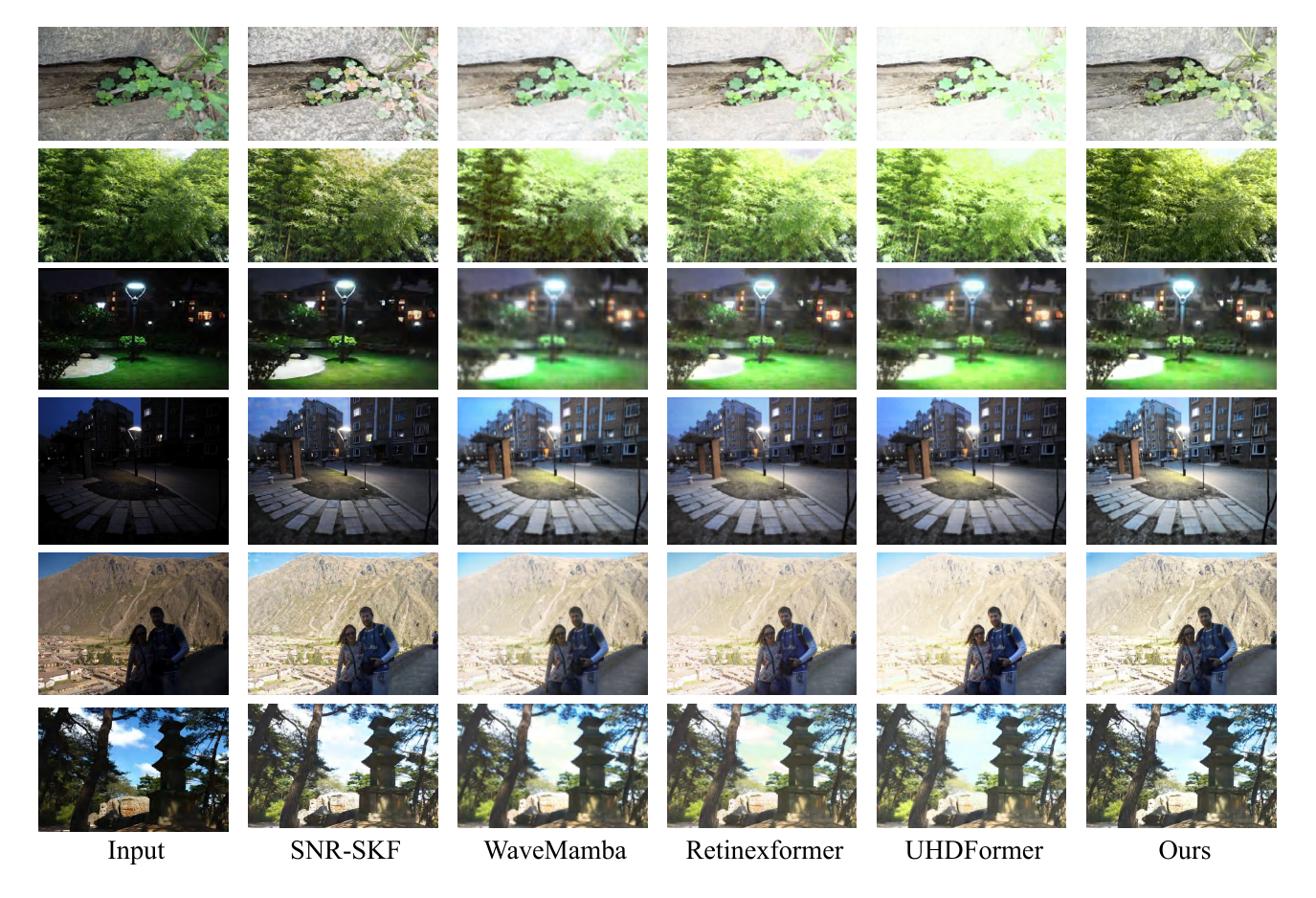}
    \caption{Visual comparison on no-reference datasets.}
    \label{fig:small_supp}
\end{figure*}

In Fig.~\ref{fig:small_supp}, we provide further visual comparisons using an unpaired dataset captured in natural scenes. This figure emphasizes our algorithm's capability to restore intricate details, color accuracy, and brightness levels, outperforming other methods significantly. The detailed comparisons illustrate the robustness of our approach in various lighting conditions, affirming its effectiveness in real-world applications.

\end{document}